\documentclass{article}
\usepackage{arxiv}
\usepackage{amsfonts,booktabs,multirow,lineno,hyperref,xcolor,amsmath}
\usepackage{blkarray, bigstrut,pifont}
\usepackage{booktabs}
\usepackage{amsthm,subfigure}
\usepackage{hyperref,booktabs}
\usepackage{algorithmic}
\usepackage{algorithm}

\usepackage[utf8]{inputenc} 
\usepackage[T1]{fontenc}    
\usepackage{hyperref}       
\usepackage{url}            
\usepackage{booktabs}       
\usepackage{amsfonts}       
\usepackage{nicefrac}       
\usepackage{microtype}      
\usepackage{lipsum}
\usepackage{graphicx}
\graphicspath{ {./images/} }

\usepackage{amsmath,amssymb,bm}
\usepackage{adjustbox}
\usepackage{caption}
\usepackage{subfigure}
\usepackage{multirow}
\usepackage[english]{babel}
\usepackage{amsthm}
\newtheorem{definition}{Definition}

\title{OWAdapt: An adaptive loss function for deep learning using OWA operators}

\author{
Sebasti\'an Maldonado\\
Department of Management Control and Information Systems \\
School of Economics and Business\\
University of Chile, Chile\\
\& Instituto Sistemas Complejos de Ingenier\'\i a (ISCI), Chile.\\
\texttt{sebastianm@fen.uchile.cl}
\And
Carla Vairetti \\
Universidad de los Andes, Chile \\
Facultad de Ingenier\'{i}a y Ciencias Aplicadas\\
\& Instituto Sistemas Complejos de Ingenier\'\i a (ISCI), Chile.\\
\texttt{cvairetti@uandes.cl}
\And
 Katherine Jara \\
 Universidad de los Andes, Chile\\
Facultad de Ingenier\'{i}a y Ciencias Aplicadas,\\
\texttt{kjara@miuandes.cl}
\And 
Miguel Carrasco \\
Universidad de los Andes, Chile\\
Facultad de Ingenier\'{i}a y Ciencias Aplicadas\\
\texttt{micarrasco@uandes.cl}
\And 
Julio L\'opez \\
Universidad Diego Portales, Chile\\
Facultad de Ingenier\'{i}a y Ciencias\\
\texttt{julio.lopez@udp.cl}
}

\begin{document}
\maketitle

\begin{abstract}
In this paper, we propose a novel adaptive loss function for enhancing deep learning performance in classification tasks. Specifically, we redefine the cross-entropy loss to effectively address class-level noise conditions, including the challenging problem of class imbalance. Our approach introduces aggregation operators to improve classification accuracy. The rationale behind our proposed method lies in the iterative up-weighting of class-level components within the loss function, focusing on those with larger errors. To achieve this, we employ the ordered weighted average (OWA) operator and combine it with an adaptive scheme for gradient-based learning. The main finding is that our method outperforms other commonly used loss functions, such as the standard cross-entropy or focal loss, across various binary and multiclass classification tasks. Furthermore, we explore the influence of hyperparameters associated with the OWA operators and propose a default configuration that performs well across different experimental settings.
\footnote{NOTICE: This is a preprint of a published work. Changes resulting from the publishing process, such as editing, corrections,structural formatting, and other quality control mechanisms may not be reflected in this version of the document. Please cite this work as follows: S. Maldonado, C. Vairetti, K. Jara, M. Carrasco, J. L\'opez (2023) OWAdapt: An adaptive loss function for deep learning using OWA operators. Knowledge-based Systems 280, 111022. DOI: \url{https://doi.org/10.1016/j.knosys.2023.111022}.}
\end{abstract}

\section{Introduction}

Deep learning (DL) has become a popular choice for various applications such as computer vision, signal processing, and text analytics \cite{dong2021survey}. The main reason for its success is its ability to learn hierarchical representations of data, which allow it to automatically extract high-level features from raw data without the need for manual feature engineering \cite{dong2021survey,lecun2015deep}.

Every DL system for supervised learning requires a loss function, which is a metric that guides the optimization process, indicating how well the model is performing \cite{lin2017focal}. Although most DL research has focused on suitable architectures for automatic feature engineering \cite{dong2021survey,lecun2015deep}, several studies have explored the design of novel loss functions \cite{heydari2019softadapt,lin2017focal,rengasamy2020deep}. In particular, adaptive loss functions have gained popularity as deep learning problems increase in sophistication. Their goal is to dynamically change the weights of a loss function to address data issues and/or to define multiple objectives \cite{barron2019general,chen2018gradnorm,huang2019addressing}.

Strategies from the literature on fuzzy systems can be used to model the uncertainty and variability in machine learning models, enhancing the capabilities of AI systems and enable them to solve more complex and challenging problems \cite{HE2022109941,Maldonado2018OWASVM,NGUYEN2020105308}. In this study, we propose the use of aggregation operators, such as the ordered weighted average (OWA), to design a generalized loss function for classification tasks using deep learning. OWA operators are very useful tools that provide a general process for reordering and aggregating the information, conferring flexibility to decision-making \cite{jin2020wa,verma2019variance}. This approach has been widely used in different domains and applications, including finance for exchange rate forecasting \cite{flores2022forecasting} and the service industry for a suitable aggregation of movie ratings \cite{CHEN2019467} or polarity detection in customer reviews \cite{serrano2022ordered}.

One relevant data issue in classification tasks is the class-imbalance problem, which arises when the label distribution is too skewed \cite{maldonado2022fw}. This is a common issue in a multiclass setting, especially when dealing with numerous labels \cite{MORTAZ2020106490}. Although resampling is the ``de facto'' approach in tabular datasets, creating new examples from under-represented classes can be difficult in DL tasks such as computer vision, text classification, or signal processing. Alternatively, cost-sensitive techniques can be adapted to DL \cite{JIN2022109817}, for example, by introducing instance-level weights in the loss function \cite{tayyar-madabushi-etal-2019-cost,zhang2016training}.

Despite the existence of some few adaptive loss functions, none of the approaches are designed to dynamically adjust class-level performance in order to guarantee the correct classification of all the classes. To overcome this research gap, we propose a novel adaptive loss function derived from traditional functions like cross-entropy loss using OWA operators. The rationale behind this approach is to adjust the weights of each class based on their performance so that the loss function can prioritize those components with greater relative error and down-weigh those with smaller relative error. This approach is particularly useful for class-imbalanced tasks where there is a shortage of data, resulting in poor class-level performance. Standard loss functions such as cross-entropy cannot resolve this issue since they assume that all classes have equal weights \cite{janocha2017loss}.

Our main contribution is the design of a fast and simple adaptive loss function that enhances the balanced performance of classifiers, bridging the research on aggregation functions and DL. Our approach is versatile and can be applied to various standard loss functions for classification. Implementing the loss function is almost effortless, and it can be incorporated into existing architectures that consider optimizers based on gradient descent approaches, such as Adam \cite{kingma2014adam}. Please note that our approach is applicable not only in scenarios that exclusively require handling class imbalance, but also in any application where a balanced classification performance in terms of class recall is desired.

The organization of this research paper is as follows: First, we review prior studies on the three pertinent topics covered in the proposal. Specifically, we discuss previous research on adaptive loss functions, class-imbalanced classification in DL, and OWA operators in Section \ref{sec:prior}. Next, in Section \ref{sec:Prop}, we present the proposed loss function for classification using DL. Then, we examine experiments conducted on benchmark datasets for image classification in Section \ref{sec:exp}. Finally, Section \ref{sec:conclusions} outlines the main findings and suggests directions for future research.

\section{Prior work}\label{sec:prior}

This section is structured as follows: First, the class-imbalance problem in deep learning tasks is discussed. Next, adaptive loss functions are formalized, and their main properties are discussed. This section concludes with the formalization of OWA operators and their relation with AI systems.

\subsection{Class-imbalance classification for deep learning tasks}\label{sec:ClassImbDL}

Several strategies have been proposed to address the class imbalance problem, which can be classified into (1) data resampling, (2) cost-sensitive learning, and (3) other algorithmic solutions \cite{Maldonado2019SMOTEFS}. Resampling can be seen as a preprocessing step in which samples from the majority class or classes are discarded. This process of downsizing the dataset is known as undersampling and can be performed either randomly or in an intelligent manner \cite{johnson2019survey}. Undersampling has been widely used in deep learning because it is the most straightforward solution to class-imbalanced classification when a relatively large sample size is available \cite{johnson2019survey}.

Alternatively, the class-imbalance issue can be tackled via data oversampling. The Synthetic Minority Over-sampling Technique (SMOTE) is arguably the most common choice in tabular datasets, which creates synthetic examples via interpolation \cite{Cha2}. However, this approach may not be suitable in most DL tasks \cite{johnson2019survey}. Some oversampling methods have been discussed in the DL literature, including the use of generative models, such as variational autoencoders \cite{dai2019generative,FAJARDO2021114463}.

Cost-sensitive techniques and other algorithmic solutions aim to deal with the class-imbalance problem during model training by taking the costs of prediction errors into account \cite{johnson2019survey}. Some strategies have been adapted to DL by e.g. introducing sample-level weights in the loss function \cite{johnson2019survey,tayyar-madabushi-etal-2019-cost}.

\subsection{Adaptive loss functions}\label{sec:loss}

The choice of the loss function in supervised classification is crucial, as it directly affects the performance and accuracy of the model. Different types of loss functions are suitable for different types of tasks, and selecting the appropriate one is important for successfully guiding the optimization process during training \cite{janocha2017loss}. 

A DL model estimates a function $f_\Theta: \textbf{x}_i \rightarrow y_i $ from data points $i=1,\ldots,N$ of the form $({\textbf x}_1,y_1),\ldots,({\textbf x}_N,y_N)$ by defining a loss function $L(f_\Theta(\textbf{x}_i), y_i)$, where $\Theta$ is the set of parameters of the model, i.e., the weights of the neural network. A multiclass problem has a target variable $\textbf{y}=(y_1,\ldots,y_N)^\top$ with $C$ classes, that is, $y_i\in\{1,2,\ldots,C\}$, for $i=1,\ldots,N$. The optimal $\Theta^*$ is obtained by minimizing the loss function:

\begin{equation}\label{loss}
\arg \min_\Theta \sum\limits_{i=1}^N L(f_\Theta(\textbf{x}_i), y_i).
\end{equation}

The class-level outputs $\hat{y}_{i,c}$ of the function $f_\Theta(\textbf{x}_i)$ can be cast into probabilities using e.g. the softmax function:
\begin{equation}\label{softmax}
p_{i,c}=\frac{\exp(\hat{y}_{i,c})}{\sum_{c=1}^{C}\exp(\hat{y}_{i,c})}.
\end{equation}
Based on these probabilities, we formalize the cross-entropy $L_{CE}$, which is arguably the best-known loss function for binary and multiclass classification \cite{nguyen2018unet}:

\begin{equation}\label{CE}
L_{CE}=-\frac{1}{N}\sum_{i=1}^{N}\sum_{c=1}^{C} y_{i,c} \log (p_{i,c}).
\end{equation}
where $y_{i,c}=1$ if $y_i=c$ and $0$ otherwise. 
A disadvantage of $L_{CE}$ is that it does not emphasize the importance of underrepresented classes. A small cross-entropy value can be achieved by focusing on samples of the majority class or classes, which usually have low misclassification costs. In order to account for the class-imbalance issue, a weighted version of the cross-entropy loss ($L_{WCE}$) follows:
\begin{equation}\label{WCE}
L_{WCE}=-\frac{1}{N}\sum_{i=1}^{N}\sum_{c=1}^{C} w_c y_{i,c} \log (p_{i,c}),
\end{equation} 
where the weights $\textbf{w}=(w_1,w_2,\ldots,w_C)$, with $w_c\in[0,1]$, can be either considered as tuning hyperparameters or computed by inverting the class frequency \cite{nguyen2018unet}. 

Although the weighted cross-entropy prioritizes underrepresented classes, its drawback is that it cannot differentiate between samples that are easy or hard to classify. To this end, the focal loss was proposed as a general version of $L_{CE}$ \cite{lin2017focal}. A tuning hyperparameter $\gamma \ge 0$ is introduced as a sample-level weighting factor for $p_{i,c}$. Focal loss is similar to $L_{WCE}$ with 
$w_c=(1 - p_{i,c})^\gamma$, and favors samples that are harder to classify: observations with low confidence in the estimated class have a small $p_{i,c}$, and they receive a weight close to one, in contrast to samples with high confidence in their estimates. 

It is important to notice that focal loss prioritizes samples that are hard to classify regardless of which class they belong to. Although this method can achieve positive results in class-imbalance classification tasks under the assumption that the underrepresented class or classes are also the ones that are harder to classify, it is not a cost-sensitive solution for the class-imbalance problem. This assumption may not hold in several cases, especially in the presence of extreme imbalances. 

Focal loss is an \emph{adaptive loss function} in the sense that it dynamically adjusts its parameters during model training according to the behavior of the learning machine \cite{heydari2019softadapt,lin2017focal}. Adaptive strategies are useful in complex settings, such as the presence of data noise or class-imbalance \cite{zhang2019deep}. 

Several adaptive loss functions have been proposed for both supervised and unsupervised learning. In \cite{zhang2019deep} proposed a novel DL approach called deep fuzzy k-means (DFKM), in which an adaptive function is used to guide the learning process. This approach is interesting because it combines fuzzy logic and adaptive loss functions in unsupervised learning. 

The Huber loss is a well-known adaptive loss function for regression that combines the $l_1$ and $l_2$ loss functions in such a way that the former is considered for large errors and the latter for small errors \cite{barron2019general}. A general version of this function was proposed in \cite{barron2019general} by introducing robustness as a continuous parameter.

SoftAdapt \cite{heydari2019softadapt} is an interesting adaptive loss function used in generative networks, such as variational autoencoders. This function dynamically updates the trade-off parameters for the various components of a multi-objective function based on the current performance at a given iteration. The authors show very good results in image reconstruction tasks. 

In the context of class-imbalanced classification, an adaptive loss function called adaptive batch nuclear-norm maximization (A-BNM) loss was proposed in \cite{Liu2023Adapt}. The rationale behind this approach is that the learning machine focuses on the majority class or classes only at the beginning of the training process. When the iterative process begins to converge, the network shifts focus to the minority class or classes, extracting the features associated with them. Similar to SoftAdapt \cite{heydari2019softadapt}, the method iteratively adapts the trade-offs of a bi-objetive function composed by an $l_2$-regularized loss function and the BNM. The latter loss function aims to maximize an upper bound of the Frobenius norm of the data matrix. Although this metric follows a similar objective to the proposed OWAdapt loss, there are important differences: First, the approach is designed to computer vision tasks under insufficient learning scenarios, such as semi-supervised learning or domain adaptation, while OWAdapt is a general-purpose metric for any classification application that requires a balanced classification performance. Secondly, our model dynamically updates class-level weights based on model performance instead of trade-off parameters of multi-part loss functions. Finally, our metric utilizes OWA operators in the adaptive learning process. 

\subsection{OWA Operators}\label{sec:OWA}

The OWA operator is arguably the best-known aggregation function from the fuzzy set theory and extends the weighted average by assigning weights to the samples based on the rank obtained when the variable is sorted. Formally, we consider an argument variable $\textbf{a}=(a_{1},a_{2},\ldots,a_{C})\in {\mathbb R}^C$, the OWA operator can be defined as follows \cite{yager1993families}:

\begin{definition} \label{def:OWA} The OWA function corresponds to a mapping $OW\!A:{\mathbb R}^{C}\rightarrow\mathbb{R}$ with a weight vector $\textbf{w}$ of dimension $C$ with $\sum_{c=1}^{C}w_{c}=1$ and $w_{c}\in [0,1]$, such that:
\begin{equation}\label{OWA}
 OW\!A(\textbf{w},\textbf{a})=\sum_{c=1}^{C}w_{c}b_{c},
\end{equation}
where $b_{c}$ is the $c$-th largest value of the $\textbf{a}$ vector.
\end{definition}

Notice that the notation used in the previous section for the classes of a classification model is considered in order to make it consistent across the study. In section \ref{sec:Prop}, the OWA operators are adapted for class-level weighting in a loss function for DL.

The OWA operator provides a set of mean-type aggregation functions between the minimum (when $w_{1} = 1$ and $w_{j}=0$ for all $c>1$) and the maximum (when $w_{c} = 1$ and $w_{c}=0$ for all $c<C$) \cite{yager1993families}. The arithmetic mean is a particular case of the OWA function, which can be obtained by defining $w_{c}=1/C$ for all $c$.

 The IOWA operator \cite{yager1999induced} is a general version of OWA in which the order is induced by a variable different from ${\bm a}$, as follows:

\begin{definition} \label{def:IOWA} The IOWA function corresponds to a mapping $IOW\!A:{\mathbb R}^{C}\rightarrow\mathbb{R}$ with a weight vector $\textbf{w}=(w_1,w_2,\ldots,w_C)$ with $\sum_{c=1}^{C}w_{c}=1$ and $w_{c}\in [0,1]$, such that:
\begin{equation}\label{IOWA}
 IOW\!A(\textbf{w},\textbf{a},\textbf{u})=\sum_{c=1}^{C} w_{c}b_{c},
\end{equation}
where $b_{c}$ is the $a_{c}$ value of the argument variable $\textbf{a}$ that has the $c$-th largest $u_c \in \textbf{u}$, where $\textbf{u}$ is the order-inducing variable \cite{yager1999induced}.
\end{definition}

The IOWA operator can be used for sample-level weighting in machine learning. In \cite{maldonado2019iowa}, for example, we generalized the hinge loss function of Fuzzy Support Vector Machine (FSVM) using this operator. The order was induced via outlier detection strategies, such as the Local Outlier Factor (LOF) method. Alternatively, a mixed-integer programming approach was proposed by \cite{marin2022soft} to perform sample-level weighting in linear and kernel-based SVM classification.

\section{Proposed loss function for adaptive class-aware learning using OWA}\label{sec:Prop}

In this section, we propose a novel adaptive loss function for classification using neural networks. The primary objective is to address class imbalance and noise by dynamically adjusting the weights of classes based on their individual performance. To leverage complex attitudinal aggregation functions, we transform class-level loss functions, like cross-entropy, into an Ordered Weighted Averaging (OWA) function. We refer to this approach as OWAdapt loss. Additionally, we discuss the path towards more sophisticated OWAdapt functions based on multi-part loss functions.

Let us redefine a general class-weighted (GCW) loss function as:
\begin{equation}\label{GCW}
L_{GCW}=\sum_{c=1}^{C} w_c F_{c}(\textbf{y},\textbf{p}),
\end{equation}
where $\textbf{y}$ and $\textbf{p}$ are the class labels and probability estimates, respectively. For example, $F_{c}(\textbf{y},\textbf{p})=-\frac{1}{N}\sum_{i=1}^{N} y_{i,c} \log (p_{i,c})$ corresponds to the cross-entropy function. Using the OWA operator formalized in Eq. \eqref{OWA}, we define the OWAdapt loss as follows:
\begin{equation}\label{OWAdapt}
L_{OW\!A}= OW\!A(\textbf{w},\textbf{F}(\textbf{y},\textbf{p}))=\sum_{c=1}^{C} w_c \hat{F}_{c}(\textbf{y},\textbf{p})
\end{equation}
where $\textbf{F}(\textbf{y},\textbf{p})=(F_1(\textbf{y},\textbf{p}),\ldots, F_C(\textbf{y},\textbf{p}))$ and 
$\hat{F}_{c}(\textbf{y},\textbf{p})$ is the $c$-th largest value of the set of class-level losses $\textbf{F}(\textbf{y},\textbf{p})$. 

The crucial component of this loss function is the set of weights $\textbf{w}$. Instead of assigning arbitrary weights, we utilize linguistic quantifiers (LQ) \cite{MALDONADO2023109058,yager2011recent}. A common choice is the regular increasing monotone (RIM) quantifier $Q$, which calculates the weights using the following expression:
\begin{equation}\label{Eqn1}
w_c=Q\left(\frac{c}{C}\right) -Q\left(\frac{c-1}{C}\right), ~~\ \mbox{for }~ c=1,\ldots,C,
\end{equation}
where $\sum_{c=1}^{C}w_{c}=1$ and $w_{c}\in [0,1]$. A function $Q$ that satisfies this requirement is the so-called basic LQ ($Q(r)=r^{\alpha}, \alpha>0$). Alternatively, popular alternatives are the quadratic LQ ($Q_{q}(r)=\left(\frac{1}{1-\alpha(r)^{0.5}}\right), \alpha>0$), or the exponential LQ ($Q_{e}(r)=e^{-\alpha r}, \alpha>0$) \cite{MALDONADO2023109058}, with $\alpha$ representing the attitudinal character of the aggregation. Notice that the weights of the quadratic and exponential LQs need to be rescaled in such a way that $\sum_{c=1}^{C}w_{c}=1$. For the exponential LQ, the antonym of $Q_{e}(r)$, denoted $\hat{Q}_{e}(r)$, is used instead of $Q_{e}(r)$ to avoid negative weights, with $\hat{Q}_{e}(r)=Q_{e}(1-r)$.

To address the issue of class imbalance, we propose an adaptive loss function that iteratively updates the weights, giving higher weightage to classes with larger errors. This prioritizes learning the classes that are more challenging to classify correctly, often coinciding with the underrepresented classes in the data. As per Eq. \eqref{OWAdapt}, the smallest weights are applied to the components with the smallest error in the loss function. In each iteration of the learning algorithm, the OWA weights are sorted based on $\textbf{F}(\textbf{y},\textbf{p})$, making the OWAdapt loss adaptive in nature.

Next, we present a high-level pseudocode for utilizing OWAdapt loss in Algorithm \ref{alg1}. 

\begin{algorithm}[!htb]
\renewcommand{\algorithmicrequire}{\textbf{Input:}}
\renewcommand{\algorithmicensure}{\textbf{Output:}}
\caption{Learning process with the OWAdapt loss function.}
\begin{algorithmic}[1]
\REQUIRE Training set $\{\textbf{X},\textbf{y}\}$, Learning rate $\eta$, Initial set of network parameters $\Theta_0$, OWA LQ hyperparameter $\alpha$. 
\STATE $\textbf{w} \ \leftarrow$ Compute the OWA weights using a quantifier function and hyperparameter $\alpha$.\\
\STATE $\Theta \ \leftarrow$ $\Theta_0$ (Network initialization).\\
\emph{Forward pass:}
 \STATE $\hat{\textbf{y}} \ \leftarrow$ Compute the predicted scores using $\textbf{X}$ and $\Theta$.
\STATE $\textbf{p} \ \leftarrow$ Compute the softmax of $\hat{\textbf{y}}$ to get the predicted probabilities using Eq. \eqref{softmax}.
\STATE $\textbf{F}(\textbf{y},\textbf{p}) \ \leftarrow$ Compute the class-level loss by relating $\textbf{y}$ and $\textbf{p}$.
\STATE $L_{OW\!A} \ \leftarrow$ Compute the OWAdapt loss using Eq. \eqref{OWAdapt} by sorting and assigning the OWA weights to the corresponding $\textbf{F}(\textbf{y},\textbf{p})$ values.\\
\emph{Backward pass and weight update:} 
\STATE $\frac{\partial L_{OW\!A}}{\partial W} \ \leftarrow$ Compute the gradient of $L_{OW\!A}$ with respect to $\textbf{W}$ using the chain rule.
\STATE $\textbf{W} := \textbf{W} - \eta \frac{\partial L_{OW\!A}}{\partial W} \ \leftarrow$ Update the weights $\textbf{W}$ using the gradient and the learning rate $\eta$.
\STATE Repeat from step 2 until a stopping criterion is fulfilled.
 \ENSURE The set of updated network parameters $\Theta$.
\end{algorithmic}
 \label{alg1}
\end{algorithm}

Note that the actual implementation may vary based on specific details of the neural network architecture and the weight update process. For instance, stochastic gradient descent algorithms like ADAM \cite{kingma2014adam} utilize mini-batches for learning instead of the entire dataset. Consequently, the size of inputs $\textbf{y}$ and $\textbf{p}$ in Step 4 corresponds to the size of the mini-batches. For example, in the case of cross-entropy, we take $F_{c}(\textbf{y},\textbf{p})=-\frac{1}{m}\sum_{i=1}^{m} y_{i,c} \log (p_{i,c})$, with $m$ denoting the number of samples in the batch. Similarly, more advanced update rules can be considered, including adaptive strategies for learning rate and momentum as used in ADAM. We provide a simple update rule with only a learning rate for illustrative purposes.

Regarding the gradient of $L_{OW\!A}$, it is straightforward to compute when a known base loss function is used. In fact, we note that 
$$
L_{OW\!A}=\sum_{c=1}^{C} w_c \hat{F}_{c}(\textbf{y},\textbf{p})=\sum_{c=1}^{C} \hat{w}_c F_c(\textbf{y},\textbf{p}),
$$
where $\hat{w}_c$ denotes the value in $\textbf{w}$ which corresponds to the $c$-th largest value of the set of class-level losses $\textbf{F}(\textbf{y},\textbf{p}).$ Therefore, the derivatives of the weighted function are a weighted sum of the derivatives of the base loss function, which are known and easy to compute. 

The proposed approach remains relevant even in the absence of extreme class imbalance. Our objective is to achieve satisfactory classification performance for all classes, as some may be inherently harder to classify regardless of label distribution. Therefore, we emphasize the worst-class recall in our experimental setting, along with other balanced performance metrics that summarize the overall classifier performance.

\section{Experimental results}\label{sec:exp}

In this study, we evaluated the proposed loss function on a total of 13 benchmark datasets. The experimental setting and relevant metadata are provided in Section \ref{sec:expSET}. The results obtained for the various loss functions considered in this study are summarized in Section \ref{sec:expRES}, followed by a sensitivity analysis for the relevant hyperparameters associated with the OWAdapt function in Section \ref{sec:sens}.

\subsection{Experimental setting}\label{sec:expSET}

We evaluated the proposed loss function on well-known object classification datasets, which offer a wide range of alternatives in terms of sample size, number of classes, and label distribution. Table \ref{tabDesc} provides a summary of the relevant information for the various datasets, including the number of samples, the number of classes, and the percentage of samples in the minority (\% Min.) and majority (\% Maj.) classes. The table is sorted according to \% Min.

\begin{table}[ht!]
\centering
\caption{Descriptive information for all benchmark datasets (D1 to D13).}
\label{tabDesc}
\begin{tabular}{cccccc}
\toprule
 ID & Name & Samples & Classes & \% Min. & \% Maj. \\
\midrule
 D1 & Cats vs dogs & 3000 & 2 & 50\% & 50\% \\
 D2 & Malaria cell & 27,558 & 2 & 50\% & 50\% \\
 D3 & Weather & 1123 & 4 & 19.06\% & 31.79\% \\
 D4 & Intel & 17,034 & 6 & 15.43\% & 17.83\% \\
 D5 & Retina oct2017 & 84,452 & 4 & 10.49\% & 44.34\% \\
 D6 & Cifar10 & 60,000 & 10 & 10\% & 10\% \\
 D7 & Fashion mnist & 70,000 & 10 & 10\% & 10\% \\
 D8 & Art images & 8577 & 5 & 9.80\% & 26.47\% \\
 D9 & Mnist & 70,000 & 10 & 9.02\% & 11.25\% \\
 D10 & QuickDrawAnimals & 14,399 & 12 & 8.33\% & 8.33\% \\
 D11 & Oregon wildlife & 13,951 & 20 & 4.12\% & 5.47\% \\
 D12 & Plant village & 61,486 & 39 & 1.63\% & 8.96\% \\
 D13 & Food 101 & 101,000 & 101 & 0.99\% & 0.99\% \\
\bottomrule
\end{tabular} 
\end{table}

Table \ref{tabDesc} provides a comprehensive overview of the 13 benchmark datasets used in our experimental evaluation. The datasets exhibit a wide range of alternatives in terms of sample size, number of classes, and label distribution. For instance, while some datasets are perfectly balanced, they still suffer from underrepresented classes because there are multiple classes available (see, e.g., D6, D7, D10, and D13). On the other hand, some datasets exhibit a skewed label distribution (see, e.g., D5, D8, D11, and D12).
 
Our experiments used three popular deep learning architectures as classifiers, including Residual Network (ResNet) \cite{resnet}, MobileNet \cite{mobilenets}, and Visual Geometry Group (VGG) \cite{vgg}. ResNet is a popular CNN variant that uses residual connections, which are useful for dealing with the vanishing gradient problem. MobileNet is a lightweight CNN model that uses depth-wise separable convolutions, reducing the number of parameters and therefore the computational cost. Finally, VGG is a very deep network that uses small filters (3x3) and is known for its simplicity and excellent performance in object detection tasks.

As alternatives to the proposed OWAdapt loss function, we also considered the standard cross-entropy (Cross-E.) and focal loss (Focal L.) in our experiments. To evaluate the average performance of a classifier, we used accuracy and F1 (macro) measures. The F1 for a given class $c$ is the harmonic mean of the precision and the recall: $F1_c = 2P_c R_c \mathbin{/}(P_c+R_c)$, where $P_c$ corresponds to the true positives divided by the sum of all positives and $R_c$ corresponds to the true positives divided by the true positives and false negatives. F1 (macro) is the mean of all the $F1_c$. F1 (macro) is a better metric than accuracy in the case of class-imbalance and was therefore chosen as the main performance metric for hyperparameter selection. 

Additionally, we reported the minimum class recall (Min. recall) and the minimum class-level F1 (Min. F1) as they illustrate the worst class-level performance. The adaptive strategy's goal is to up-weight the loss components associated with the worst class-level performances in the training set. Hence, we expected to confirm that OWAdapt could provide the best-balanced performance, ensuring a good minimum performance for the underrepresented classes.

For hyperparameter selection, we used cross-validation within the training process. For the OWAdapt loss, we implemented the basic, quadratic, and exponential linguistic quantifiers with $\alpha \in \{0.2,0.4,0.5,0.6,0.7,0.8,0.9,0.99\}$, providing a wide range of weights for the adaptive process. These values were previously considered in the context of machine learning (see \cite{MALDONADO2023109058}). For the focal loss, we considered $\gamma \in \{0,1,2,5\}$ following \cite{lin2017focal}.

As for the CNN classifiers, we used default configurations of pre-trained models from the `torchvision.models' PyTorch sub-package (resnet50, mobilenet-v2, and vgg19-bn). We used a stochastic gradient descent (SGD) optimizer with a learning rate of 0.003 and a momentum of 0.9, and five epochs as the stopping criterion (see, e.g., \cite{bjorck2018understanding}). We considered fixed parameters for the optimizer in order to isolate the effect of the various loss functions in the learning process. Experiments with fixed class-level cost relations were not performed because we were not able to find suitable benchmark datasets with this information. 

\subsection{Results summary}\label{sec:expRES}

The detailed results for the 13 datasets, four performance metrics, three CNN classifiers, and three loss functions are presented in Appendix A. Our experiments show that our loss function performs better, in general, than cross-entropy and focal loss. Although we obtain the best performance in most comparisons for the four relevant evaluation metrics (accuracy, F1, minimum recall and minimum F1), there are some cases in which the alternative loss functions perform better than OWAdapt. For example, focal loss achieves best F1 on the D2 and D13 datasets when ResNet is used as the classifier, D4 and D13 datasets when MobileNet is used as the classifier, and D7 and D13 datasets when VGG is used as the classifier.

To summarize the results, we rank each loss function based on its relative position across all datasets and CNN classifiers. A ranking of 1 means that the method was the top-performing one for a given dataset and CNN classifier. Next, we average all the ranks to obtain the overall performance for each loss function. The second and third columns of Table \ref{tabTest} show the average rank (A. Rank) and the average performance on a given measure (A. Metric), respectively. The proposed OWAdapt loss achieves the best overall rank (minimum value in the second column) and performance in the corresponding metric (maximum value in the third column).

To assess the statistical significance of the best model's performance, we implemented the Friedman test with Iman-Davenport correction, which was proposed in \cite{Demsar2006} for this purpose in the context of machine learning. This test computes a chi-squared statistic for each of the four performance metrics, assessing whether there are significant differences between the various loss functions. The Friedman statistics with Iman-Davenport correction were 72.72414, 81.17113, 98.24555, and 157.3316 for accuracy, F1 (macro), minimum class recall, and minimum F1, respectively, rejecting the null hypothesis of similar performances between the loss functions with a p-value below 0.000001. These results confirm that the OWAdapt loss achieves significantly better performance than other alternatives, and the impact is larger in a balanced metric such as F1 (macro) in relation to accuracy. Additionally, the statistical differences are more extreme in the metrics designed to assess the worst class-level performance.

The Holm test was designed to assess whether or not the top-ranked technique outperforms others statistically \cite{Demsar2006}. Once the average ranks are computed, Nemenyi's Z test is estimated for pairwise comparisons between the top method (OWAdapt loss in the four cases) and the remaining ones. The outcome of this test is 'reject' when its p-value is below a threshold $\beta/(j-1)$, with $k=2,3$ being the overall rank for an alternative loss function and $\beta=5\%$ the significance level. The results for the Holm test are presented in the last three columns of Table \ref{tabTest}. This table includes the p-value of the test (fourth column), the threshold (fifth column), and the outcome (sixth column).

\begin{table}[ht!]
\caption{Holm test for pairwise comparisons for the three loss functions (OWAdapt, focal loss, and cross-entropy) obtained using all 13 datasets.}
\label{tabTest}
\centering
\begin{tabular}{lccccc}
\toprule
Method & A. Rank & A. Metric & p-value & $\frac{\beta}{(j-1)}$ & Outcome \\
\midrule
\multicolumn{ 4}{l}{\emph{Accuracy as the performance measure}} & & \\
 OWAdapt & 1.231 & 76.793 & - & - & - \\
 Focal L. & 1.923 & 75.147 & 0.002 & 0.05 & reject \\
 Cross-E. & 2.846 & 70.531 & 0.000 & 0.025 & reject \\
\multicolumn{ 4}{l}{\emph{F1(macro) as the performance measure}} & & \\
 OWAdapt  & 1.231 & 76.200 & - & - & - \\
 Focal L. & 1.897 & 74.485 & 0.003 & 0.05 & reject \\
 Cross-E. & 2.872 & 69.474 & 0.000 & 0.025 & reject \\
\multicolumn{ 4}{l}{\emph{min. recall as the performance measure}} & & \\
 OWAdapt  & 1.167 & 67.997 & - & - & - \\
 Focal L. & 1.974 & 64.171 & 0.000 & 0.05 & reject \\
 Cross-E. & 2.859 & 57.143 & 0.000 & 0.025 & reject \\
\multicolumn{ 4}{l}{\emph{min. F1 as the performance measure}} & & \\
 OWAdapt  & 1.192 & 67.182 & - & - & - \\\
 Focal L. & 1.846 & 63.626 & 0.004 & 0.05 & reject \\
 Cross-E. & 2.962 & 55.548 & 0.000 & 0.025 & reject \\
\bottomrule
\end{tabular}
\end{table}

As shown in Table \ref{tabTest}, OWAdapt is able to statistically outperform both focal loss and cross-entropy for all four metrics, making it an excellent alternative for classification tasks with underrepresented labels. Focal loss performs better on average than cross-entropy, showing that flexible loss functions based on cross-entropy can achieve superior performance than this strategy.

\subsection{Sensitivity analysis}\label{sec:sens}

To discuss the influence of the hyperparameters associated with the OWAdapt loss, we report the best configurations obtained for this function on the 13 datasets using F1 (macro) in Table \ref{tabSens}. The results are presented for the three CNN classifiers ResNet (R), MobileNet (M), and VGG (V). The relevant hyperparameters are the $\alpha$ values and the three linguistic quantifiers: Basic (B), Quadratic (Q), and Exponential (E).

\begin{table}[ht!]
\caption{Best hyperparameter configuration for the OWAdapt loss function in terms of the linguistic quantifier (LQ): Basic (B), Quadratic (Q), and Exponential (E), and the quantifier parameter $\alpha \in \{0.2,0.4,0.5,0.6,0.7,0.8,0.9,0.99\}$ The results are presented for each dataset DS (D1 to D13) and CNN classifier: ResNet (R), MobileNet (M), and VGG (V).}
\label{tabSens}
\centering
\begin{tabular}{cccccccc}
\toprule
 DS & CNN & LQ & $\alpha$ & DS & CNN & LQ & $\alpha$ \\
\midrule
\multicolumn{ 1}{c}{D1} & R & E & 0.2 & \multicolumn{ 1}{c}{D8} & R & Q & 0.5 \\
\multicolumn{ 1}{c}{} & M & B & 0.8 & \multicolumn{ 1}{c}{} & M & E & 0.6 \\
\multicolumn{ 1}{c}{} & V & Q & 0.5 & \multicolumn{ 1}{c}{} & V & E & 0.8 \\
\multicolumn{ 1}{c}{D2} & R & E & 0.8 & \multicolumn{ 1}{c}{D9} & R & E & 0.2 \\
\multicolumn{ 1}{c}{} & M & E & 0.9 & \multicolumn{ 1}{c}{} & M & E & 0.2 \\
\multicolumn{ 1}{c}{} & V & Q & 0.9 & \multicolumn{ 1}{c}{} & V & E & 0.7 \\
\multicolumn{ 1}{c}{D3} & R & E & 0.6 & \multicolumn{ 1}{c}{D10} & R & E & 0.6 \\
\multicolumn{ 1}{c}{} & M & E & 0.4 & \multicolumn{ 1}{c}{} & M & E & 0.9 \\
\multicolumn{ 1}{c}{} & V & E & 0.99 & \multicolumn{ 1}{c}{} & V & E & 0.5 \\
\multicolumn{ 1}{c}{D4} & R & E & 0.99 & \multicolumn{ 1}{c}{D11} & R & E & 0.8 \\
\multicolumn{ 1}{c}{} & M & E & 0.6 & \multicolumn{ 1}{c}{} & M & E & 0.5 \\
\multicolumn{ 1}{c}{} & V & E & 0.8 & \multicolumn{ 1}{c}{} & V & E & 0.9 \\
\multicolumn{ 1}{c}{D5} & R & E & 0.5 & \multicolumn{ 1}{c}{D12} & R & E & 0.9 \\
\multicolumn{ 1}{c}{} & M & E & 0.99 & \multicolumn{ 1}{c}{} & M & E & 0.8 \\
\multicolumn{ 1}{c}{} & V & E & 0.7 & \multicolumn{ 1}{c}{} & V & E & 0.99 \\
\multicolumn{ 1}{c}{D6} & R & E & 0.8 & \multicolumn{ 1}{c}{D13} & R & B & 0.9 \\
\multicolumn{ 1}{c}{} & M & E & 0.9 & \multicolumn{ 1}{c}{} & M & Q & 0.7 \\
\multicolumn{ 1}{c}{} & V & E & 0.9 & \multicolumn{ 1}{c}{} & V & B & 0.99 \\
\multicolumn{ 1}{c}{D7} & R & E & 0.99 & & & & \\
\multicolumn{ 1}{c}{} & M & E & 0.99 & & & & \\
\multicolumn{ 1}{c}{} & V & E & 0.9 & & & & \\
\bottomrule
\end{tabular}
\end{table}

Table \ref{tabSens} shows that the exponential quantifier (LQ=E) achieves the best performance in approximately 80\% of the cases. Therefore, we recommend using this linguistic quantifier as a default configuration. As for the quantifier parameter, the best performance is obtained with $\alpha \geq 0.8$ in about 60\% of the cases. Hence, we recommend exploring large values for this hyperparameter.

To evaluate the stability of the results, we consider two illustrative datasets. First, D10 does not exhibit class-imbalance, but the fact that it has 12 classes with 8.33\% of the samples each can lead to asymmetrical classification performance. Second, D12 shows signs of class imbalance, with 39 classes and sizes ranging from 1.63\% to 8.96\%.

\begin{figure}[ht!]
\centering
\subfigure[D10 Dataset, ResNet]{%
 \label{Fig1:first}
 \includegraphics[width=0.31\textwidth]{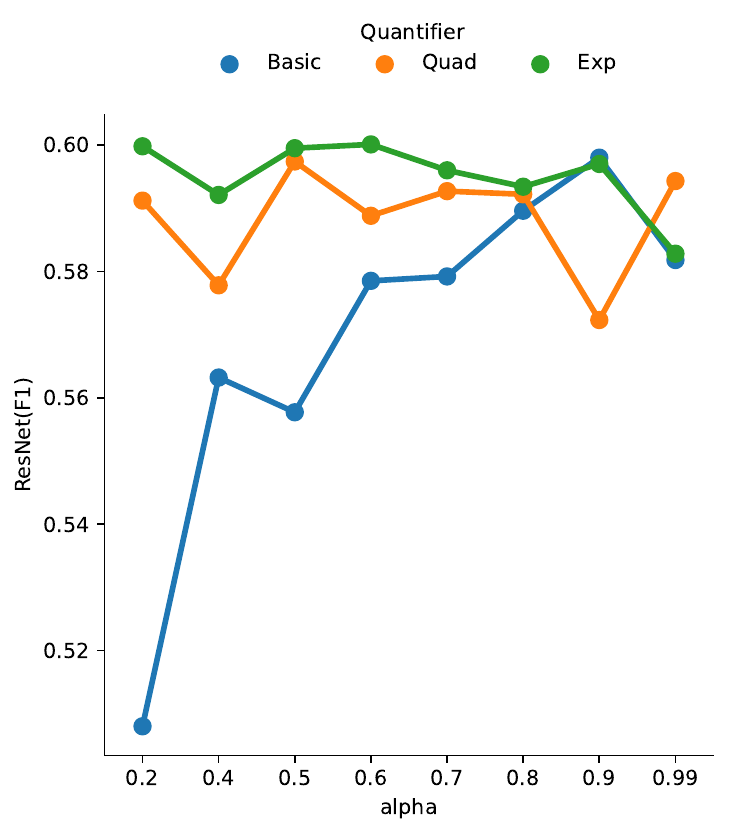}
 }%
 \subfigure[D10 Dataset, MobileNet]{%
 \label{Fig1:second}
 \includegraphics[width=0.31\textwidth]{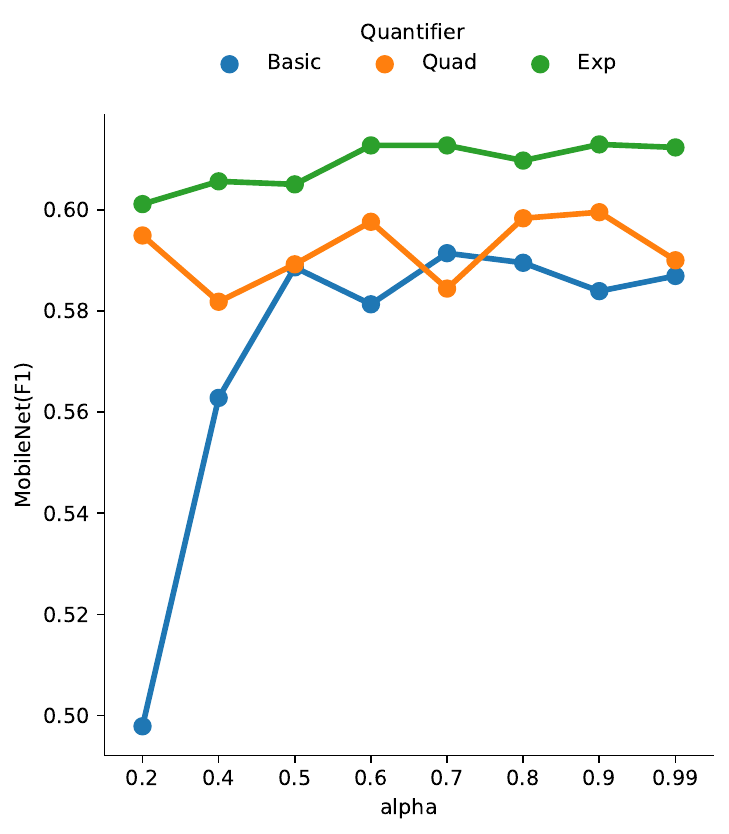}
  }%
\subfigure[D10 Dataset, VGG]{%
 \label{Fig1:third}
 \includegraphics[width=0.31\textwidth]{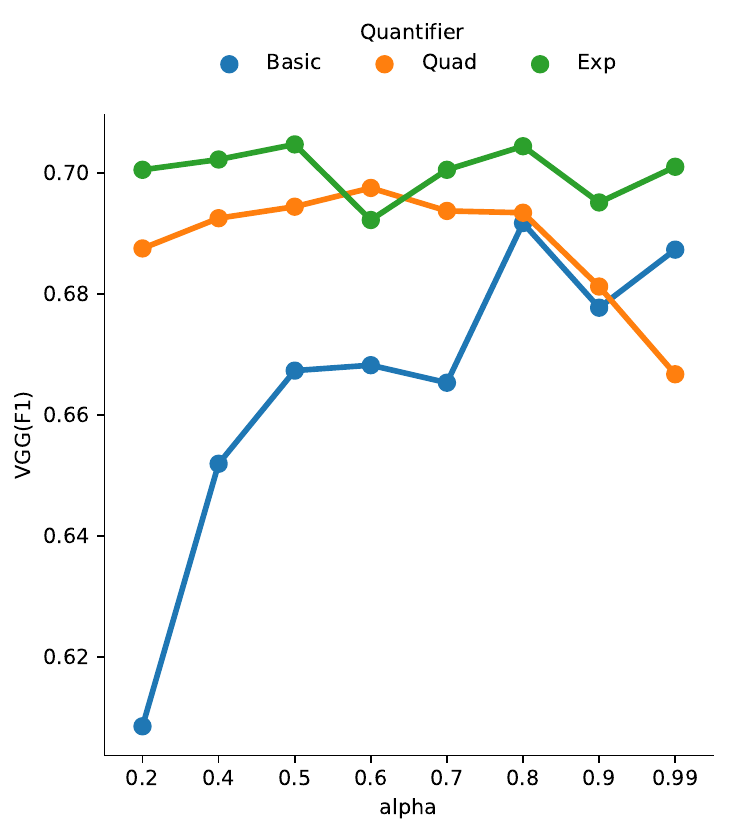}
}\\ 
 \subfigure[D12 Dataset, ResNet]{%
 \label{Fig1:fourth}
 \includegraphics[width=0.31\textwidth]{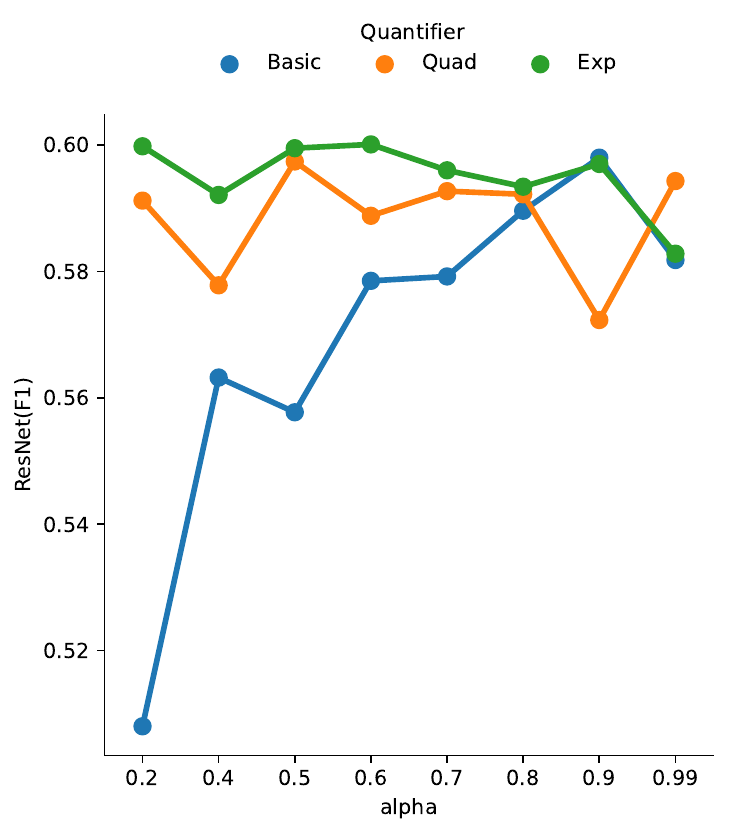}
 }
 \subfigure[D12 Dataset, MobileNet]{%
 \label{Fig1:fifth}
 \includegraphics[width=0.31\textwidth]{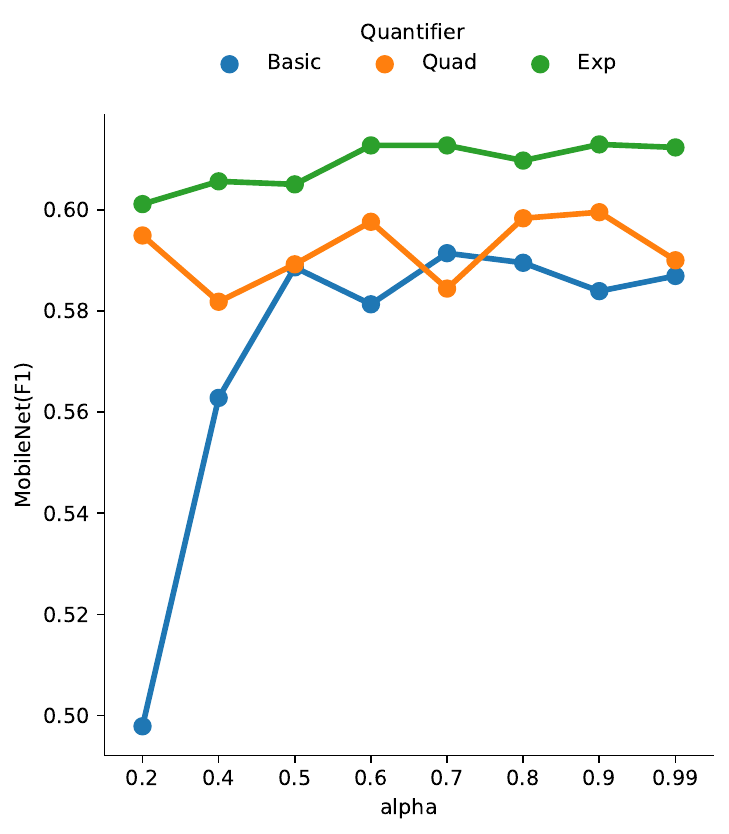}
 }%
 \subfigure[D12 Dataset, VGG]{%
 \label{Fig1:sixth}
 \includegraphics[width=0.31\textwidth]{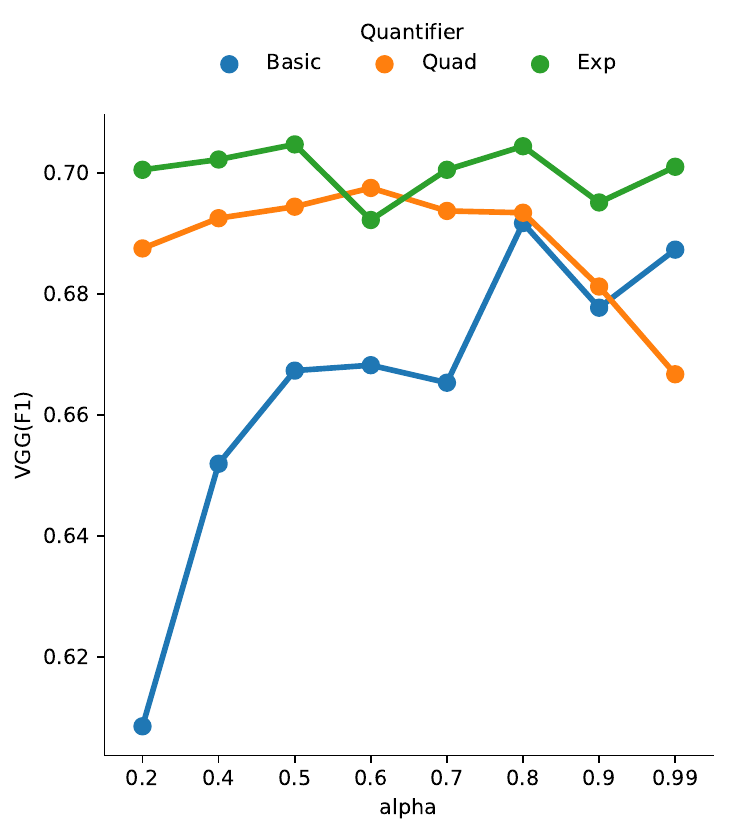}
 }\\ 
 \caption{Classification performance of OWadapt for the various hyperparameter values on two datasets (D10 and D12). F1 (macro) as the performance measure.}%
 \label{Fig:1}
\end{figure}

Figure \ref{Fig:1} shows the F1 (macro) for an increasing $\alpha$ value and the three quantifiers. The information is presented in six parts: Figures \ref{Fig1:first}, \ref{Fig1:second}, and \ref{Fig1:third} illustrate the performance for D10 using the Resnet, MobileNet, and VGG classifiers, respectively. Figures \ref{Fig1:fourth}, \ref{Fig1:fifth}, and \ref{Fig1:sixth} present the same information for D12.

Figure \ref{Fig:1} indicates that the quadratic and exponential quantifiers exhibit very stable classification performance for the different $\alpha$ values (green and orange lines). In contrast, the basic linguistic quantifier (blue lines) performs poorly for small $\alpha$ values. We conclude that the exponential quantifier suggested above provides robust performance for different $\alpha$ values. However, we recommend tuning this hyperparameter carefully to account for the variability observed in the datasets.

It is important to note that there are more OWA quantifiers than the ones reported in this study. For example, we considered the trigonometric quantifier in other machine learning studies (see, e.g., \cite{Maldonado2018OWASVM,maldonado2022fw}). However, we observed better results with the basic RIM quantifier. Alternatively, the O'Hagan method is another popular approach that defines a mathematical programming problem with the objective of maximizing a dispersion measure \cite{Luuka}.

\section{Discussion and conclusions}\label{sec:conclusions}

\subsection{Strengths and limitations}

This study presents a novel adaptive loss function for classification via OWA operators. The main virtue of this function is its general and adaptive structure that can prioritize the classes that are harder to classify in the iterative learning process of the algorithm. The reasoning behind this approach is directly linked to the class-imbalance problem, as underrepresented labels are usually the ones that are harder to classify. In many applications, misclassifying underrepresented classes has higher costs than other classes, which can negatively affect the decision-making process (see, e.g., \cite{Maldonado2020ProfMPM,Zhang2019}). On the other hand, the proposed method extends the existing scientific literature on fuzzy systems, as this is the first study to incorporate OWA operators in the optimization process behind neural networks, to the best of our knowledge.

Experiments were carried out on benchmark datasets, showing the strengths of the OWAdapt loss function in terms of performance and stability. The proposed function outperformed cross-entropy and focal loss statistically, achieving excellent overall and class-level performance on 13 datasets. Finally, a sensitivity analysis was performed, concluding that a default configuration using the exponential linguistic quantifier is possible.

One limitation of the proposed loss function is that it is not able to account for sample-level noise. In contrast to other adaptive loss functions that use sample-level weights, we consider a class-level prioritization scheme. A sample-level strategy could be useful to address label noise, for example. The presence of unreliable categories of the target variable can severely degrade the generalization ability of deep learning models, especially when they are intentionally manipulated by users \cite{Song2022}. Alternatively, labels can be corrupted because the ground truth can be complex even for specialists, which is common in medical domains \cite{Song2022}. We consider the use of aggregation functions to deal with this issue as future work.

A limitation of OWAdapt and focal loss is that they have additional parameters to be tuned, in contrast to the traditional cross-entropy loss. The latter is recommended for its simplicity in traditional DL classification when the distribution of the labels is relatively balanced and/or the misclassification costs are similar.  

OWA operators offer a versatile tool in multi-criteria decision-making; however, they have some limitations, which were evidenced in this study. First, the results are highly dependent on the choice of weights, and there is no straightforward method to determine the best set of weights for a given problem. In our study, we guided the process by the classification performance of the method. Additionally, interpretability is an issue. The implications of various weight vectors can be non-intuitive, especially for non-experts \cite{jin2020wa,Luuka}. Despite these limitations, the proposed OWAdapt loss resulted in a very effective and stable method for flexible deep learning.
  
\subsection{Future work: cost-sensitive OWAdapt based on the OWAWA operator}

As an interesting avenue for future research, the OWAdapt loss can be extended as a multi-part adaptive function for cost-sensitive learning by incorporating a fixed cost vector and the OWAWA operator. This approach involves two weighting processes: the WA and the OWA. The former establishes a weighted average based on predefined class relationships, while the latter performs the adaptive mechanism proposed for the OWAdapt loss.

The OWA operator can be extended further to combine different objectives, which can be useful in the definition of loss functions. In \cite{MALDONADO202197}, a novel FSVM method was proposed to deal with noisy examples and outliers and dataset shift, which is the issue that occurs when the joint distribution of inputs and outputs evolves over time. The weighted average (WA) operator was used to weigh down potentially outdated samples. This is one of the simplest forms of aggregation function, and has the following form \cite{merigo2011unified}:

\begin{definition} \label{def:WA} The WA function corresponds to a mapping $WA:{\mathbb R}^{C}\rightarrow{\mathbb R}$ with a weight vector $\textbf{w}=(w_1,w_2,\ldots,w_C)$ with $\sum_{c=1}^{C}w_{c}=1$ and $w_{c}\in [0,1]$, such that:
\begin{equation}\label{WA}
 W\!A(\textbf{a})=\sum_{c=1}^{C}w_{c}a_{c}.
\end{equation}
\end{definition}

The combination between the OWA operator with the weighted average leads to the OWAWA function, as follows:

\begin{definition} \label{def:OWAWA} The OWAWA operator corresponds to a mapping $OW\!AW\!A:\mathbb{R}^{C}\rightarrow\mathbb{R}$ with two weight vectors $\textbf{w}$ and $\textbf{v}$ of size $C$ with $\sum_{c=1}^{C}w_{c}=\sum_{c=1}^{C}v_{c}=1$ and $w_{c},\ v_{c}\in [0,1]$, such that:
\begin{equation}\label{OWAWA}
OW\!AW\!A(\textbf{w},\textbf{v},\textbf{a})=\sum_{c=1}^{C} \hat{v}_{c}b_{c}
\end{equation}
where $b_{c}$ is the $c$-th largest value of the argument variable $\textbf{a}$. Vector $\hat{\textbf{v}}$ can be expressed as $\hat{v}_{c}=\beta w_c + (1-\beta) v_c$, where $v_c$ is the weight $v_{c'}$ ordered according to $\textbf{b}$. Parameter $\beta \in [0,1]$ provides a trade-off between the WA and OWA operators: $\beta=1$ leads to the OWA operator, while $\beta=0$ to the WA operator \cite{merigo2011unified}.
\end{definition}

In order to define a cost-sensitive version of the OWAdapt loss function, we consider two weight vectors $\textbf{w}$ and $\textbf{v}$ of size $C$ for the OWA and WA components, respectively, with $\sum_{c=1}^{C}w_{c}=\sum_{c=1}^{C}v_{c}=1$ and $w_{c},\ v_{c}\in [0,1]$. The OWAdapt loss with fixed class-level cost relations is defined as:
\begin{align}\label{OWAWAdapt}
L_{OW\!AW\!A}&= OW\!AW\!A(\textbf{w},\textbf{v},{\bf F}
(\textbf{y},\textbf{p}))\\
&=\sum_{c=1}^{C} \hat{v}_c \hat{F}_{c}(\textbf{y},\textbf{p})
\end{align}
where $\hat{F}_{c}(\textbf{y},\textbf{p})$ represents the $c$-th largest value from the set of class-level losses ${\bf F}(\textbf{y},\textbf{p})$. Similar to Definition \ref{def:OWAWA}, $\hat{\textbf{v}}$ is a two-part weighting vector that uses $\beta \in [0,1]$ as a trade-off parameter. For each class $c$, $\hat{v}_{c}=\beta w_c + (1-\beta) v_c$, where $v_c$ is the weight $v_{c'}$ ordered according to $\hat{\bf F}(\textbf{y},\textbf{p})$. The weight vector $\textbf{w}$ can be obtained using OWA quantifiers, while $\textbf{v}$ represents a set of normalized relations between labels. For instance, consider a seven-class task where the first three classes have a 2:1 cost relation with the remaining classes. In this case, $\textbf{v}=\{0.2,0.2,0.2,0.1,0.1,0.1,0.1\}$. Consequently, the OWAWA operator will up-weight the first three classes during each iteration of the learning algorithm (WA operator), while the OWA mechanism will up-weight the classes with larger errors.

A completely different experimental setting is required to evaluate the $L_{OW\!AW\!A}$ because we need to establish cost relations between classes. This is a common approach in business analytics and other domains \cite{Maldonado2020ProfMPM,Zhang2019} but is less frequent in computer vision, which represents the current experimental framework for this study. Therefore, an empirical evaluation of this metric is proposed as future work.

There are some additional avenues for future development that could be explored. Another source of noise is dataset shift, as indicated in Section \ref{sec:OWA}. Our weighting strategy can be used to address this issue. Additionally, OWAdapt can be easily extended to other DL tasks, such as multilabel classification.

\bibliographystyle{IEEEtran}
\bibliography{biblio-OWA-loss}

\clearpage
\appendix
\renewcommand{\theequation}{\thesection.\arabic{equation}} 
\setcounter{equation}{0} 
\renewcommand{\thetable}{\thesection.\arabic{table}} 
\setcounter{table}{0} 
\section{Detailed Results for all datasets}\label{sec:detaileda}

Tables \ref{tabPerf1} to \ref{tabPerf2} present the results for all the datasets and loss functions using accuracy and F1, and minimum class recall and minimum F1, respectively. The best performance for each dataset and metric is emphasized in bold type. 

\begin{table}[ht!]
\centering
\caption{Detailed results for all the benchmark datasets. Accuracy and F1 as performance metrics.}
\label{tabPerf1}
\begin{tabular}{ccccccc}
\toprule
 & \multicolumn{ 3}{c}{Accuracy} & \multicolumn{ 3}{c}{F1} \\
 & OWA & CE & FL & OWA & CE & FL \\
\midrule
\multicolumn{ 3}{l}{\emph{ResNet as the DL classifier}} & & & & \\
 D1 & {\bf 65.8} & 61.3 & 61.1 & {\bf 65.58} & 59.99 & 61.09 \\
 D2 & {\bf 92.33} & 90.8 & 92.43 & 92.33 & 90.8 & {\bf 92.43} \\
 D3 & {\bf 70.97} & 67.9 & 70 & {\bf 71.18} & 68.32 & 70.2 \\
 D4 & {\bf 71.2} & 68.47 & 68.67 & {\bf 71.3} & 68.33 & 68.83 \\
 D5 & {\bf 96.38} & 80.17 & 94.21 & {\bf 96.39} & 78.68 & 94.18 \\
 D6 & {\bf 73.56} & 67.48 & 69.65 & {\bf 73.43} & 67.27 & 69.65 \\
 D7 & {\bf 91.77} & 90.21 & 91.44 & {\bf 91.75} & 90.28 & 91.44 \\
 D8 & {\bf 77.22} & 72.78 & 74.07 & {\bf 72.3} & 67.55 & 68.89 \\
 D9 & {\bf 99.39} & 98.97 & 99.21 & {\bf 99.39} & 98.96 & 99.2 \\
 D10 & {\bf 60.23} & 58.07 & 59.02 & {\bf 60.01} & 57.99 & 58.9 \\
 D11 & {\bf 79.86} & 55.25 & 77.89 & {\bf 79.84} & 54.91 & 77.85 \\
 D12 & {\bf 86.23} & 79.59 & 84.07 & {\bf 83.69} & 76.04 & 81.47 \\
 D13 & 7.59 & 13.46 & {\bf 14.1} & 6.94 & 11.84 & {\bf 13.88} \\
\multicolumn{ 3}{l}{\emph{MobileNet as the DL classifier}} & & & & \\
 D1 & {\bf 66.8} & 62.9 & 63.6 & {\bf 66.78} & 62.58 & 63.6 \\
 D2 & {\bf 93.65} & 92.16 & 92.83 & {\bf 93.65} & 92.16 & 92.83 \\
 D3 & {\bf 73.83} & 71.57 & 72.8 & {\bf 74.07} & 71.88 & 73 \\
 D4 & 72.93 & 71.87 & {\bf 73.37} & 73.06 & 71.99 & {\bf 73.61} \\
 D5 & {\bf 94.32} & 70.25 & 80.89 & {\bf 94.28} & 67.11 & 79.83 \\
 D6 & {\bf 74.27} & 69.61 & 71.02 & {\bf 74.12} & 69.77 & 70.83 \\
 D7 & {\bf 92.09} & 90.57 & 91.14 & {\bf 92.09} & 90.5 & 91.14 \\
 D8 & {\bf 77.92} & 75.47 & 74.77 & {\bf 74.07} & 70.65 & 68.98 \\
 D9 & {\bf 99.44} & 98.87 & 99.23 & {\bf 99.43} & 98.87 & 99.23 \\
 D10 & {\bf 61.57} & 55.73 & 58.94 & {\bf 61.29} & 54.9 & 58.85 \\
 D11 & {\bf 77.1} & 39.77 & 70.33 & {\bf 77.16} & 38.44 & 70.3 \\
 D12 & {\bf 86.67} & 77.83 & 83.07 & {\bf 84.2} & 74.29 & 80.25 \\
 D13 & 7.55 & 13.68 & {\bf 16.86} & 6.37 & 12.15 & {\bf 15.93} \\
\multicolumn{ 3}{l}{\emph{VGG as the DL classifier}} & & & & \\
 D1 & {\bf 74.4} & 68.6 & 67.9 & {\bf 74.39} & 68.46 & 67.9 \\
 D2 & {\bf 95.16} & 94.58 & 94.74 & {\bf 95.16} & 94.57 & 94.74 \\
 D3 & {\bf 79.73} & 77.63 & 79.13 & {\bf 80.14} & 77.87 & 79.5 \\
 D4 & {\bf 80.57} & 77.67 & 79.23 & {\bf 80.78} & 77.98 & 79.48 \\
 D5 & {\bf 97.73} & 85.12 & 94.73 & {\bf 97.71} & 84.07 & 94.71 \\
 D6 & {\bf 83.31} & 78.78 & 82.63 & {\bf 83.39} & 78.79 & 82.69 \\
 D7 & 93.62 & 92.91 & {\bf 93.66} & 93.58 & 92.85 & {\bf 93.65} \\
 D8 & {\bf 80.96} & 77.34 & 78.62 & {\bf 76} & 71.98 & 73.28 \\
 D9 & {\bf 99.55} & 99.11 & 99.52 & {\bf 99.55} & 99.11 & 99.52 \\
 D10 & {\bf 70.61} & 66.53 & 68.7 & {\bf 70.47} & 66.3 & 68.68 \\
 D11 & {\bf 82.34} & 38.41 & 78.9 & {\bf 82.31} & 36.78 & 78.79 \\
 D12 & {\bf 89.97} & 82.01 & 88.03 & {\bf 87.93} & 78.61 & 85.73 \\
 D13 & 16.3 & 17.27 & {\bf 20.23} & 15.68 & 15.87 & {\bf 19.85} \\
\bottomrule
\end{tabular} 
\end{table}

\begin{table}[ht!]
\centering
\caption{Detailed results for all the benchmark datasets. Minimum class recall and F1 as performance metrics.}
\label{tabPerf2}
\begin{tabular}{ccccccc}
\toprule
 & \multicolumn{ 3}{c}{Min. Recall} & \multicolumn{ 3}{c}{Min. F1} \\
 & OWA & CE & FL & OWA & CE & FL \\
\midrule
\multicolumn{ 3}{l}{\emph{ResNet as the DL classifier}} & & & & \\
 D1 & {\bf 65.15} & 58.3 & 60.69 & {\bf 65.13} & 52.75 & 60.35 \\
 D2 & {\bf 92.14} & 89.65 & 91.55 & 92.26 & 90.74 & {\bf 92.41} \\
 D3 & 65.03 & 54.17 & {\bf 65.53} & {\bf 64.39} & 62.95 & 63.36 \\
 D4 & {\bf 65.68} & 63.69 & 61.65 & {\bf 65.96} & 61.61 & 62.22 \\
 D5 & {\bf 93.08} & 68.75 & 88.64 & {\bf 95.46} & 60.23 & 92.44 \\
 D6 & {\bf 57.63} & 49.2 & 54.46 & {\bf 55.79} & 49.05 & 49.85 \\
 D7 & {\bf 79.09} & 69.58 & 77.56 & 76.51 & 74.25 & {\bf 76.62} \\
 D8 & {\bf 58.65} & 50 & 53.78 & {\bf 53.98} & 48.78 & 52.29 \\
 D9 & {\bf 99.02} & 98.49 & 98.84 & {\bf 99.13} & 98.36 & 98.86 \\
 D10 & {\bf 37.42} & 26.22 & 33.67 & {\bf 36.67} & 30.86 & 33.82 \\
 D11 & {\bf 73.83} & 37.4 & 65 & {\bf 72.82} & 34.51 & 69.57 \\
 D12 & {\bf 70.67} & 51.74 & 64.22 & {\bf 66.85} & 52.26 & 64.16 \\
 D13 & 1.12 & 0 & {\bf 3.1} & 0.93 & 0 & {\bf 3.2} \\
\multicolumn{ 3}{l}{\emph{MobileNet as the DL classifier}} & & & & \\
 D1 & {\bf 65.97} & 60.88 & 63.44 & {\bf 65.91} & 59.1 & 63.38 \\
 D2 & {\bf 93.35} & 91.71 & 92.37 & {\bf 93.6} & 92.06 & 92.82 \\
 D3 & {\bf 67.68} & 64.92 & 66.42 & {\bf 68.24} & 65.41 & 66.53 \\
 D4 & {\bf 67.47} & 64.94 & 66.96 & 66.67 & 65.29 & {\bf 66.9} \\
 D5 & {\bf 90.64} & 62.69 & 72.42 & {\bf 93.28} & 43.21 & 65.22 \\
 D6 & {\bf 58.49} & 47.99 & 56.14 & {\bf 55.83} & 49.07 & 52.55 \\
 D7 & {\bf 81.54} & 77.39 & 78.69 & {\bf 78.25} & 72.45 & 76.45 \\
 D8 & {\bf 62.12} & 56.16 & 50 & {\bf 59.75} & 52.23 & 46.72 \\
 D9 & {\bf 99.23} & 97.92 & 98.88 & {\bf 99.27} & 98.07 & 98.91 \\
 D10 & {\bf 41.38} & 26.67 & 33.57 & {\bf 39.82} & 29.09 & 30.43 \\
 D11 & {\bf 68.97} & 22.22 & 59.14 & {\bf 70.42} & 15.76 & 58.04 \\
 D12 & {\bf 71.86} & 49.55 & 61.86 & {\bf 67.22} & 48.82 & 61.86 \\
 D13 & 0 & 0 & {\bf 4} & 0 & 0 & {\bf 1.33} \\
\multicolumn{ 3}{l}{\emph{VGG as the DL classifier}} & & & & \\
 D1 & {\bf 73.53} & 66.43 & 67.51 & {\bf 73.82} & 66.38 & 67.54 \\
 D2 & {\bf 94.75} & 92.95 & 93.9 & {\bf 95.14} & 94.51 & 94.73 \\
 D3 & {\bf 71.77} & 69.76 & 69.81 & {\bf 73.54} & 70.53 & 71.17 \\
 D4 & {\bf 74.87} & 68.67 & 71.91 & {\bf 74.62} & 69.26 & 71.61 \\
 D5 & {\bf 95.65} & 75.16 & 91.54 & {\bf 96.6} & 67.75 & 93.97 \\
 D6 & {\bf 69.28} & 59.02 & 69.02 & {\bf 68.64} & 60.85 & 66.24 \\
 D7 & {\bf 84.65} & 83.14 & 83.63 & 81.05 & 77.74 & {\bf 81.38} \\
 D8 & 60.56 & {\bf 60.71} & 56.79 & {\bf 56.29} & 52.48 & 54.01 \\
 D9 & 99.21 & 97.37 & {\bf 99.22} & {\bf 99.27} & 98.23 & 99.16 \\
 D10 & {\bf 47.54} & 42.11 & 39.13 & {\bf 45.43} & 36.36 & 41.86 \\
 D11 & {\bf 74.45} & 15 & 67.83 & {\bf 75.79} & 10.46 & 68.25 \\
 D12 & {\bf 74.73} & 55.71 & 65.1 & {\bf 72.34} & 54.24 & 67.15 \\
 D13 & 3.7 & 2.33 & {\bf 4.69} & 3.43 & 0.68 & {\bf 4.05} \\
\bottomrule
\end{tabular} 
\end{table}

\end{document}